\documentclass[conference]{IEEEtran}
\IEEEoverridecommandlockouts
\usepackage{cite}
\usepackage{amsmath,amssymb,amsfonts}
\usepackage{algorithmic}
\usepackage{graphicx}
\usepackage{textcomp}
\usepackage{xcolor}
\def\BibTeX{{\rm B\kern-.05em{\sc i\kern-.025em b}\kern-.08em
    T\kern-.1667em\lower.7ex\hbox{E}\kern-.125emX}}
\begin{document}

\title{MACeIP: A Multimodal Ambient Context-enriched Intelligence Platform in Smart Cities}

\author{\IEEEauthorblockN{
Truong Thanh Hung Nguyen\IEEEauthorrefmark{1},
Phuc Truong Loc Nguyen\IEEEauthorrefmark{1},
Monica Wachowicz\IEEEauthorrefmark{2},
Hung Cao\IEEEauthorrefmark{1}}
\IEEEauthorblockA{
\IEEEauthorrefmark{1}Analytics Everywhere Lab, University of New Brunswick, Canada,\\
\IEEEauthorrefmark{2}RMIT University, Australia\\
 Email: \{hung.ntt, hcao3\}@unb.ca, monica.wachowicz@rmit.edu.au 
 }
 }

\maketitle

\begin{abstract}
This paper presents a Multimodal Ambient Context-enriched Intelligence Platform (MACeIP) for Smart Cities, a comprehensive system designed to enhance urban management and citizen engagement. Our platform integrates advanced technologies, including Internet of Things (IoT) sensors, edge and cloud computing, and Multimodal AI, to create a responsive and intelligent urban ecosystem. Key components include Interactive Hubs for citizen interaction, an extensive IoT sensor network, intelligent public asset management, a pedestrian monitoring system, a City Planning Portal, and a Cloud Computing System. We demonstrate the prototype of MACeIP in several cities, focusing on Fredericton, New Brunswick. This work contributes to innovative city development by offering a scalable, efficient, and user-centric approach to urban intelligence and management.
\end{abstract}


\section{Introduction}
Rapid urbanization and technological advancements have given rise to the concept of intelligent cities, leveraging digital technologies to enhance quality of life, improve operational efficiency, and promote sustainable development.
As cities grow more complex, there is an increasing need for integrated systems to collect, process, and utilize vast amounts of data to inform decision-making and improve urban services.
Hence, in this paper, we contribute as follows:
\begin{enumerate}
    \item We detail the architecture of our Multimodal Ambient Context-enriched Intelligence Platform (MACeIP) in Smart Cities, including Interactive Hubs for citizen engagement, an IoT sensor network, intelligent public asset management, an advanced pedestrian monitoring system, a City Planning Portal (CPP), and a Cloud Computing System (CCS) equipped by Multimodal AI.
    \item Our Multimodal AI system includes time-series and vision models, Large Language Models (LLMs), and Explainable AI (XAI) techniques to deliver a citizen-centric, intelligent, efficient, and responsible decision-making platform for smart cities.
    \item We describe MACeIP deployment in Canadian cities, focusing on Fredericton, New Brunswick. We also demonstrate the user interfaces of the Interactive Hubs and CPP.
\end{enumerate}

\section{Related Work}
\begin{figure}[ht!]
    \centering
    \includegraphics[width=0.7\linewidth]{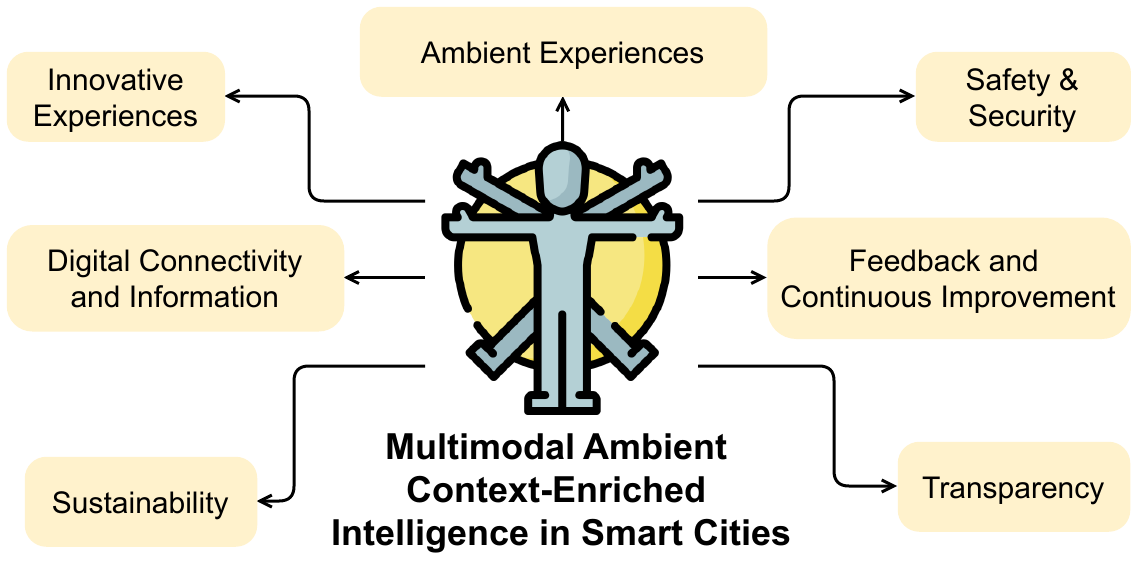}
    \caption{Factors of MACeIP in Smart Cities}
    \label{fig:factors}
\end{figure}

The development of MACeIP in Smart Cities encompasses several interconnected research areas, as outlined in our conceptual framework (Figure \ref{fig:factors}).
This section explores the literature related to key components identified in our framework.

Ambient and innovative experiences in smart cities have gained significant attention, with \cite{streitz2019grand} exploring ambient intelligence in urban environments and \cite{kakderi2019smart} discussing innovative digital services. \cite{bibri2017smart} have extensively studied sustainability in smart cities, highlighting the integration of information and communication technology with urban systems for environmental, economic, and social sustainability.
\cite{laufs2020security} addressed safety and security challenges in smart cities, emphasizing the need for integrated security systems. 
Transparency and digital connectivity, crucial for building trust and engagement, have been explored by \cite{berntzen2016role}, underscoring the importance of open data policies and accessible digital infrastructure.
Feedback and continuous improvement, central to adaptive smart cities, have been examined \cite{anthopoulos2017understanding}, emphasizing iterative development and continuous learning in urban innovation. Recent research has also highlighted the potential of multimodal approaches in enhancing urban intelligence, with \cite{shin2022multimodal} proposing a surveillance map data framework and \cite{raj2023improved} emphasizing the integration of heterogeneous data for comprehensive urban analysis.

% \cite{shin2022multimodal} proposes a surveillance map data framework for integrated multimodal data representation from various agents (e.g., Closed-circuit television (CCTV), delivery robots). This framework supports anomaly detection for security services by analyzing environmental data collected by multiple agents.
% \cite{raj2023improved} emphasizes the integration of heterogeneous data from sensors, social media, and IoT devices for comprehensive urban analysis. This approach enables the detection of complex patterns, such as social behavior and traffic congestion, using advanced machine learning (ML) techniques.
% \subsection{Ambient Intelligence}

% \subsection{Multimodal Context-enriched Approaches for Smart Cities}

\section{Orchestration}
This section presents a detailed description of the orchestration of our MACeIP for Smart Cities. The system integrates multiple technologies and components to create an intelligent, responsive, responsible urban management ecosystem.

\begin{figure}[ht!]
    \centering
    \includegraphics[width=\linewidth]{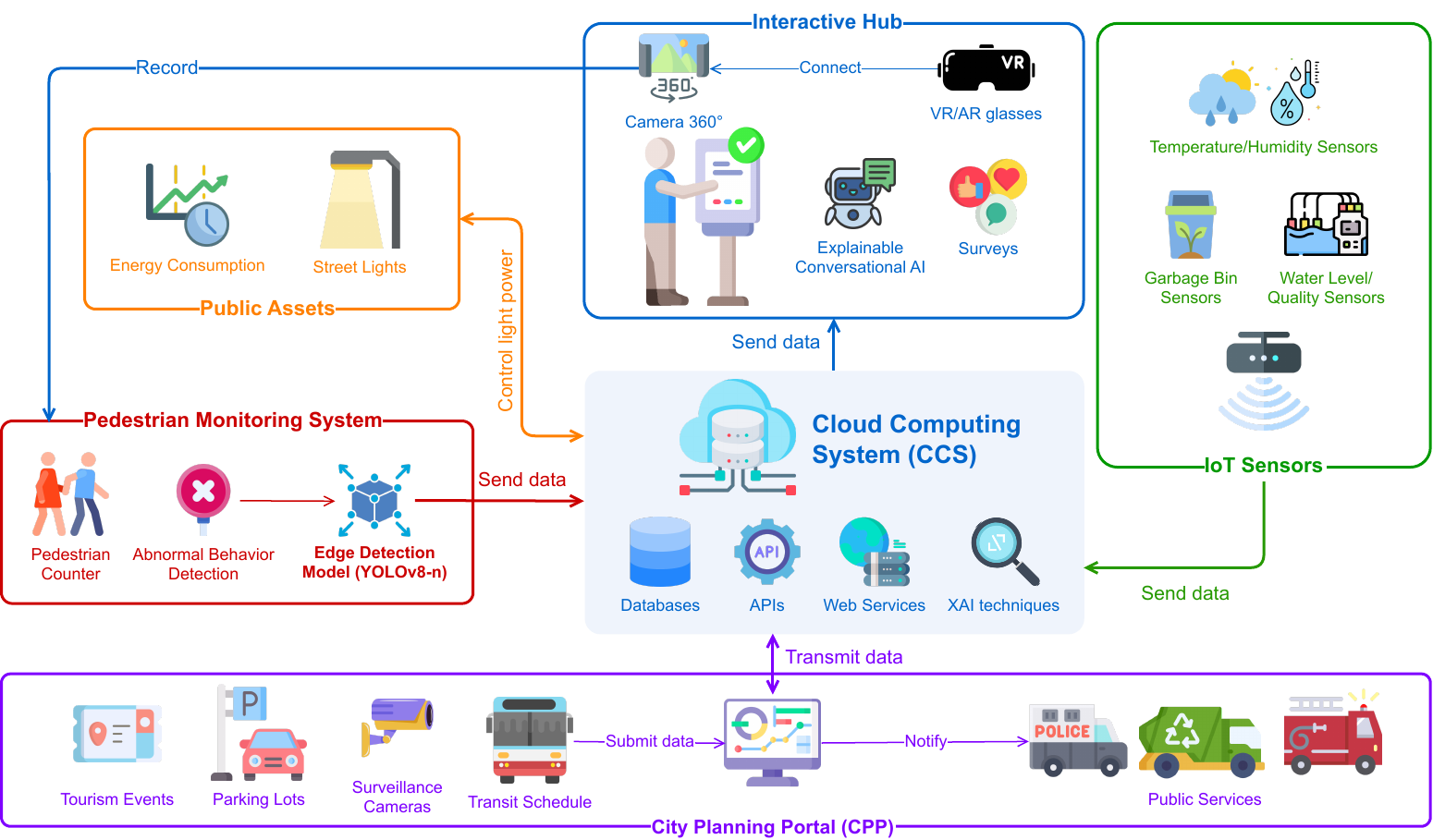}
    \caption{Orchestration of Multimodal Ambient Context-enriched Intelligence Platform (MACeIP) in Smart Cities}
    \label{fig:enter-label}
\end{figure}

\subsection{Interactive Hub}\label{ss:ih}
The Interactive Hub serves as the primary interface between citizens and the smart city infrastructure, which are placed throughout public areas and act as multifunctional information and interaction points.
At the core of the Interactive Hub's functionality is an Explainable Conversational AI powered by LLMs, namely GPT-4o mini \cite{GPT4omin55:online}, and integrated with XAI techniques, such as G-CAME \cite{nguyen2024efficient}, D-CLOSE \cite{truong2023towards}, which provides transparent and understandable responses, supporting multilingual interactions for diverse urban populations. 
The user interface (UI) displays real-time weather information, transit schedules, and interactive buttons for accessing various city services. 
To enhance user experience (UX), the system incorporates VR/AR capabilities, allowing citizens to virtually explore different locations in the city through a network of 360-degree cameras.
Data collection and privacy are critical aspects of the Interactive Hub's design. 
The system implements secure data collection protocols, providing users with transparent information about data usage and privacy policies and encouraging trust in the smart city infrastructure.

\subsection{IoT Sensors Network}
The MACeIP leverages an extensive network of IoT sensors to gather real-time data about various aspects of the urban environment, which forms our data-driven approach to city management.
Our sensor array includes:
\begin{enumerate}
    \item Temperature and Humidity Sensor: We use the DHT22 device for its accuracy and reliability in monitoring ambient temperature and humidity.
    \item Water Level Sensor: We use an ultrasonic sensor, HC-SR04 \cite{morgan2014hc}, to measure water levels in reservoirs and flood-prone areas.
    \item Water Quality Sensor: We use the Analog Total Dissolved Solids (TDS) Sensor Meter for Arduino to measure TDS in water.
    \item Garbage Bin Sensor: We use HC-SR04 \cite{morgan2014hc} to measure the distance from the top of the bin to the garbage.
    \item Air Quality Sensor: We use MQ-135 Gas Sensor to monitor air quality, especially harmful gas, such as $\text{CO}_2$, $\text{NH}_3$, $\text{NO}$, $\text{NO}_2$.
    \item Ultraviolet (UV) Index Sensor: We use the VEML6070 UV sensor to measure the UV light intensity and index.
\end{enumerate}

Data transmission from these sensors relies on LoRaWAN protocol based on the Low-Power Wide-Area Network (LPWAN) technology, which ensures efficient long-range, low-power communication.
Edge computing is employed for preliminary data processing and noise filtering, reducing the load on the CCS (Section \ref{ss:CCS}).

\subsection{Pedestrian Monitoring System}
The advanced Pedestrian Monitoring System uses a network of cameras and AI-powered analytics to enhance public safety and optimize city services.
The camera network consists of 360-degree cameras installed at the Interactive Hub and traditional surveillance cameras to provide comprehensive coverage of public spaces.
The cores of this system are edge detection models based on the YOLOv8n \cite{varghese2024yolov8}, a lightweight architecture suitable for edge devices to perform real-time detection.
The detection models are designed for two major tasks: pedestrian counting and abnormal behaviour detection.
The pedestrian counting model is trained with the City Persons dataset \cite{zhang2017citypersons}, while the abnormal behaviour detection model is trained with the Fall Detection dataset \cite{fall-detection-ca3o8_dataset} and the Abnormal Behavior Detection dataset \cite{abnormal-behavior-detection_dataset}.

\subsection{Public Assets Management}
The intelligent management of public assets focuses on efficiency, sustainability, and public safety.
A key component of this system is the smart street lighting network.
Each street light is network-connected, allowing for remote monitoring and control, and includes energy consumption metering.
The energy management system built in the CCS uses historical consumption data and a time-series ML algorithm to forecast energy demand, enabling proactive management of resources.
We train the long short-term memory (LSTM) model \cite{hochreiter1997long} for forecasting time-series energy demand.
The system dynamically controls power distribution, adjusting street light intensity based on historical demands, the number of pedestrians, specific events, weather conditions, and time of day.

\subsection{City Planning Portal (CPP)}
A CPP serves as a comprehensive dashboard and management interface for urban planners, administrators, and decision-makers.
Its UI features a customizable dashboard displaying real-time city status metrics with interactive data visualizations for trend analysis.

The CPP incorporates a notification system that generates alerts based on predefined rules and urban events. The system integrates with public service communication channels, enabling seamless coordination and automated escalation procedures for critical issues between municipal departments, such as police, fire department, and emergency services.

\subsection{Cloud Computing System (CCS)}\label{ss:CCS}
The CCS forms the backbone of our smart city platform, providing the computational power, storage, and intelligence necessary for managing the entire ecosystem.
Built on Azure cloud services, the CCS contains the following components:
\begin{enumerate}
    \item Data Management: Azure Cosmos DB manages diverse data types, offering scalability and global distribution. It handles structured, unstructured, and vector data, providing a robust data storage and retrieval backbone.
    \item Application Programming Interface (API) Management: We use Azure API Management for secure, scalable API creation and management. This ensures seamless communication between different components of the smart city infrastructure and applications.
    \item Web Services: We use Azure App Services to deploy and manage web applications and services, such as CPP.
    \item Multimodal AI Management: We use Azure Machine Learning (e.g., time-series forecast models, pedestrian monitoring models) and Azure AI Studio (e.g., LLMs) with the Responsible AI dashboard to manage AI models, ensuring they are explainable and transparent, ensuring compliance with data protection regulations.
\end{enumerate}

\section{Implementation}
% Describe the conceptual implementation
Our conceptual architecture aligns with the Urgent Edge Computing architecture proposed by \cite{dazzi2024urgent}, which consists of five interconnected layers designed to support the unique requirements of time-critical, edge-based applications. This architecture, illustrated in Figure \ref{fig:conceptual-architecture}, provides a comprehensive framework for implementing our edge computing solutions. 
The five layers are as follows:
\begin{figure}[ht!]
\centering
\includegraphics[width=\linewidth]{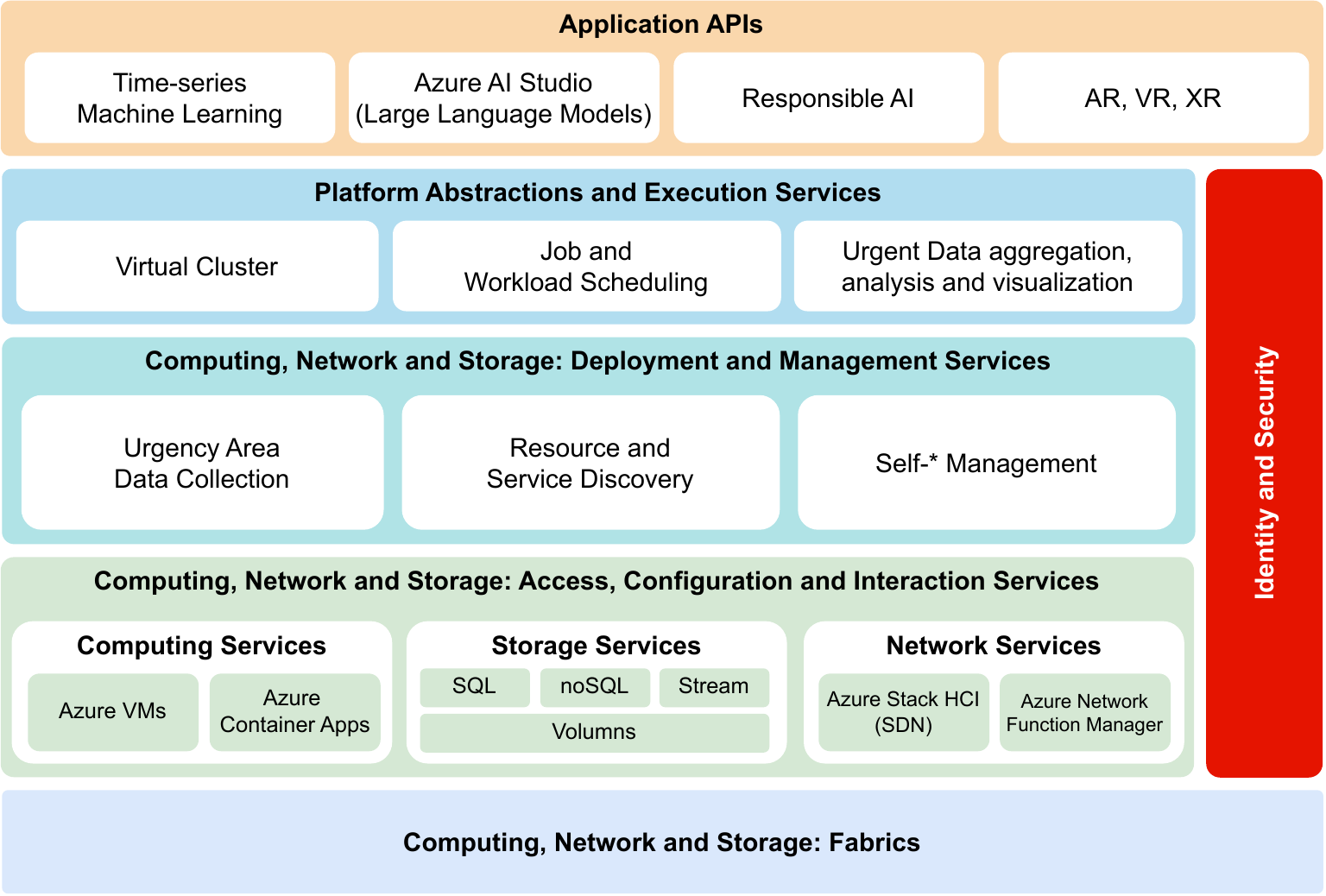}
\caption{The MACeIP edge computing conceptual architecture}
\label{fig:conceptual-architecture}
\end{figure}

\begin{enumerate}
    \item Computing, Network and Storage – Fabrics: This foundational layer represents the physical and virtual resources that form the backbone of our computing environment. It encompasses many devices and infrastructure components, from edge devices to cloud resources, that can be leveraged for urgent computing tasks.
    \item Computing, Network and Storage – Access, Configuration and Interaction Services: This layer provides the necessary interfaces and services for accessing, configuring, and interacting with the underlying computing, network, and storage resources. It includes Azure Virtual Machines (VMs), Azure Container Apps, and Azure Stack HCI for software-defined networking (SDN).
    \item Computing Network and Storage – Deployment and Management Services: This layer focuses on efficiently deploying and managing resources within our edge computing ecosystem. It incorporates critical functionalities like resource and service discovery, self-management capabilities, and urgency area data collection mechanisms.
    \item Platform Abstractions and Execution Services: This layer offers higher-level abstractions and services that facilitate the execution of computing tasks. It includes components for virtual cluster management, job and workload scheduling, and urgent data aggregation, analysis, and visualization.
    \item Application APIs: The topmost layer provides interfaces and tools for developing applications that effectively utilize the edge computing infrastructure. This includes support for Azure ML, Responsible AI and integration with Azure AI Studio to leverage LLMs. It also supports AR, VR, and XR for enhanced user interaction.
\end{enumerate}

This architecture also incorporates:
\begin{enumerate}
    \item Identity and Security: Security measures are implemented across all layers to ensure data integrity and protect sensitive information in urgent scenarios.
    \item Urgency-Aware Resource Management: The architecture incorporates mechanisms for prioritizing and allocating resources based on the urgency of computing tasks, ensuring critical operations receive the necessary computational power.
\end{enumerate}

\begin{figure}[h!]
    \centering
    \includegraphics[width=.8\linewidth]{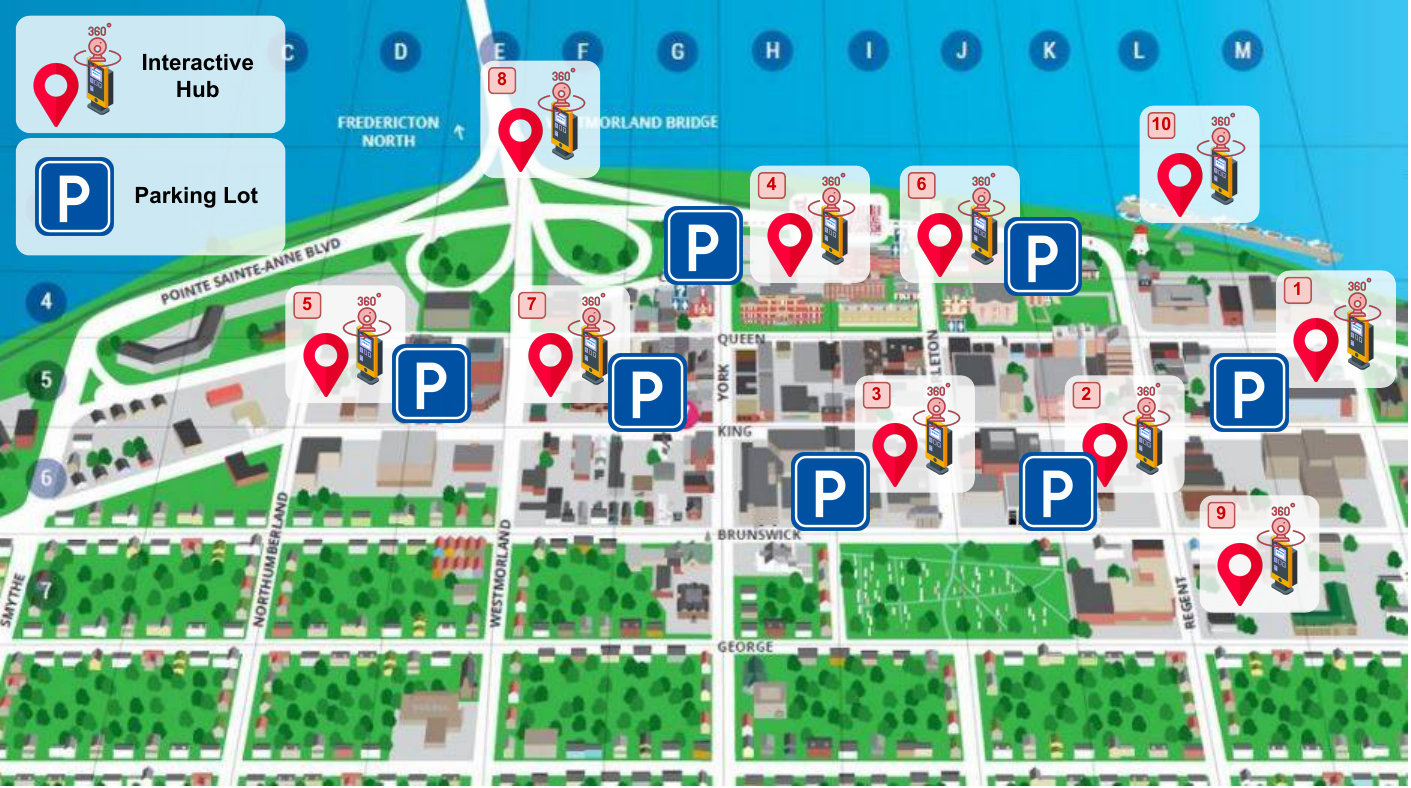}
    \caption{Our prototype installation map in Fredericton, New Brunswick, Canada}
    \label{fig:map}
\end{figure}

\begin{figure}[h!]
    \centering
    \includegraphics[width=.5\linewidth]{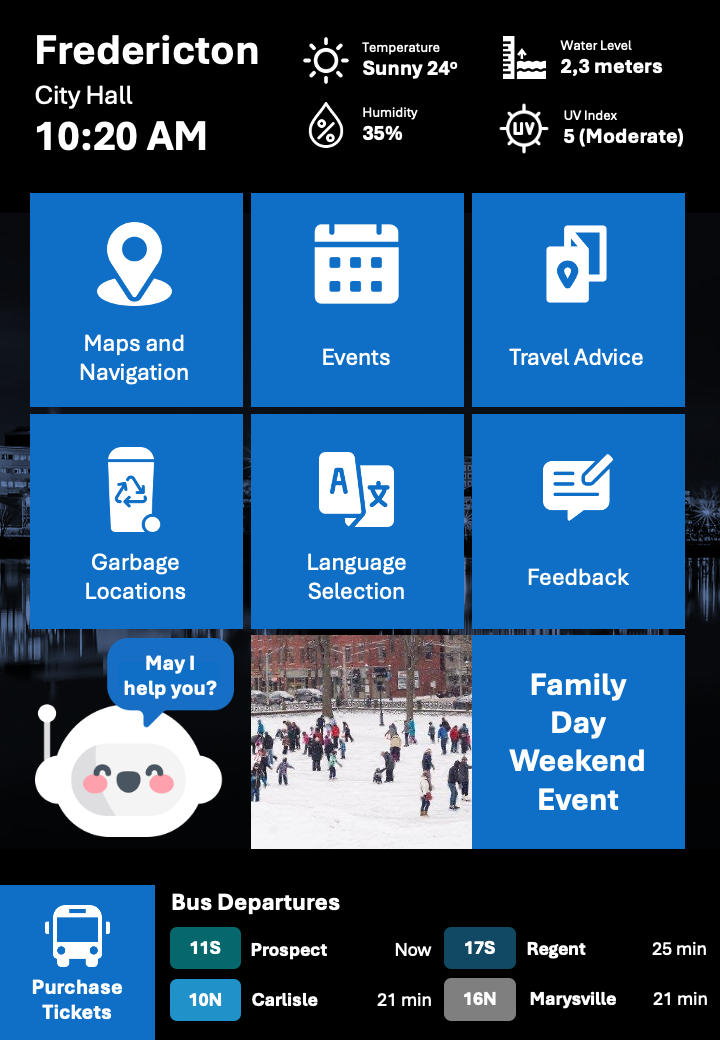}
    \caption{Interactive Hub Interface}
    \label{fig:hub}
\end{figure}

\begin{figure}[h!]
    \centering
    \includegraphics[width=.8\linewidth]{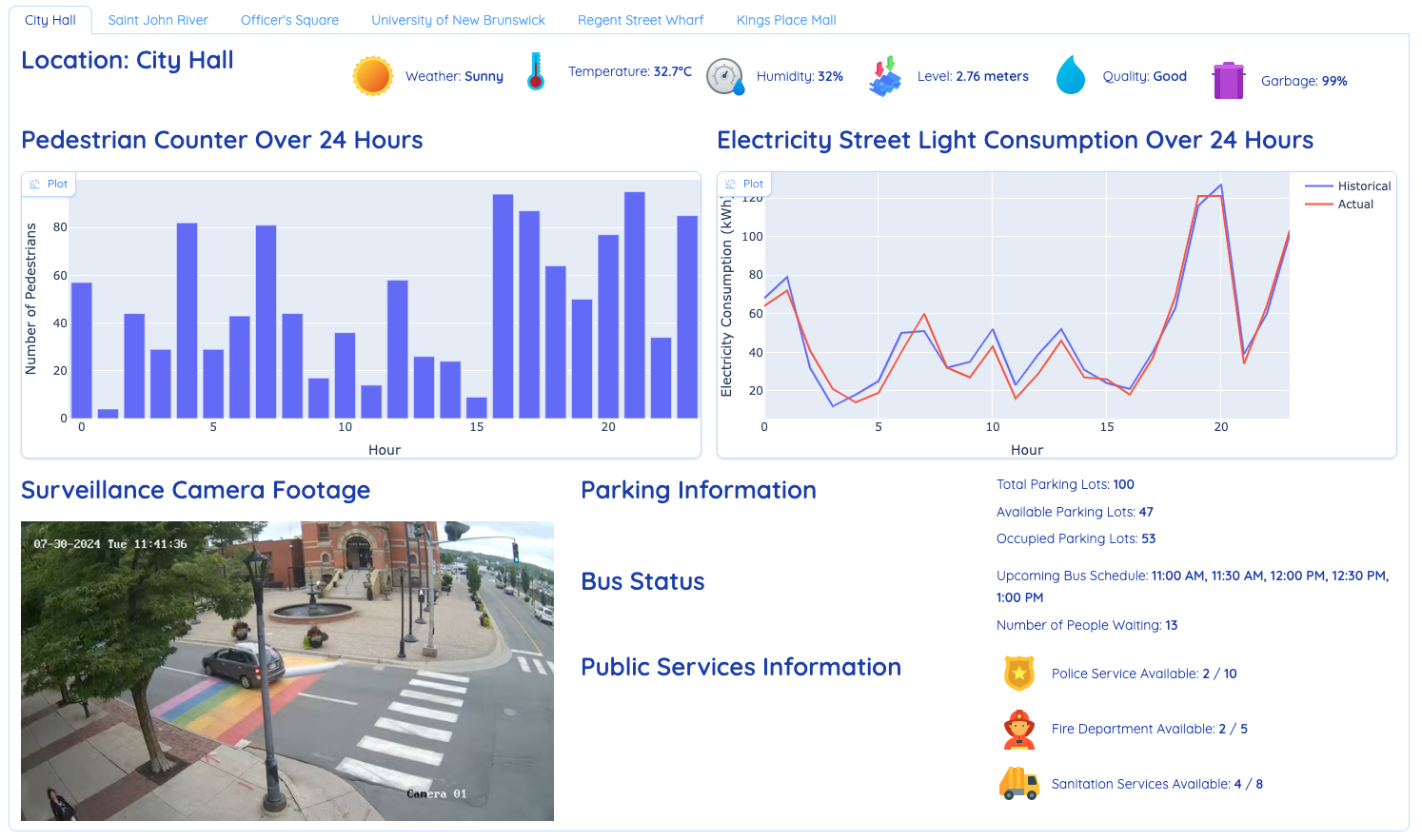}
    \caption{The City Planning Portal (CPP) User Interface}
    \label{fig:CPPI}
\end{figure}

\subsection{Installation Locations}
We demonstrate our MACeIP in several cities across Canada, with Fredericton, New Brunswick, showcased in this section.
Interactive Hubs (Section \ref{ss:ih}) are placed in 10 public locations throughout Fredericton City. 
These locations are detailed in Figure \ref{fig:map}.
Six hubs are co-located with parking lots (Numbers 1-6), while the remaining four are in public and tourist areas (Numbers 7-10).

\subsection{Tourism Hub Interface}
The Tourism Hub Interface (Figure \ref{fig:hub}) provides a comprehensive user experience catering to residents and visitors. It offers real-time information and services, including maps and navigation, event listings, travel advice, garbage locations, language selection, feedback collection, and bus schedules. Users can also purchase tickets for public transportation via this interface. Additionally, it supports interactions via an AI-powered assistant.

\subsection{CPP Interface}
The CPP Interface (Figure \ref{fig:CPPI})  is an advanced dashboard designed for urban planners, administrators, and decision-makers. It displays real-time metrics and visualizations, including pedestrian counts, streetlight electricity consumption, surveillance camera footage, parking information, bus status, and public service availability.

\section{Conclusion}
This paper presents the MACeIP, which represents an advancement in smart city technologies by integrating IoT sensors, edge computing, Multimodal AI, and cloud services, offering a comprehensive approach to urban management and citizen engagement.
Our prototype implementation demonstrates the platform's practical applicability, enabling efficient city operations and enhanced citizen services through strategically placed Interactive Hubs, an IoT sensor network, and advanced analytics capabilities. 
MACeIP provides a framework for creating intelligent, efficient, responsible, and citizen-centric urban environments.

\section*{Acknowledgments}
This work was funded by the NBIF Talent Recruitment Fund (TRF2003-001) and the UNB-FCS Startup Fund (22-23 START UP/ H CAO). The computing resource used for this study was partially supported by CFI Project Number 39473 - Smart Campus Integration and Testing (SCIT Lab).

\bibliographystyle{IEEEtran}
\bibliography{IEEEabrv,ref}
\end{document}